\documentclass{article}

\usepackage{amsmath,amsfonts,bm}

\def\eqref#1{equation~\ref{#1}}

\def\1{\bm{1}}

\DeclareMathAlphabet{\mathsfit}{\encodingdefault}{\sfdefault}{m}{sl}
\SetMathAlphabet{\mathsfit}{bold}{\encodingdefault}{\sfdefault}{bx}{n}

\newcommand{\R}{\mathbb{R}}

\newcommand{\minim}[1]{\underset{#1}{\text{min }}}
\newcommand{\maxim}[1]{\underset{#1}{\text{max }}}

\def \R { \mathbb R }

\def \bmb { \begin{bmatrix} }
\def \bme { \end{bmatrix} }

\newcommand{\cA}{{\mathcal{A}}}

\newcommand{\cD}{{\mathcal{D}}}

\newcommand{\cM}{{\mathcal{M}}}

\newcommand{\cS}{{\mathcal{S}}}

\newcommand{\cT}{{\mathcal{T}}}

\newcommand{\cX}{{\mathcal{X}}}

\usepackage{arxiv}

\usepackage[utf8]{inputenc} 
\usepackage[T1]{fontenc}    
\usepackage{hyperref}       
\usepackage{url}            
\usepackage{booktabs}       
\usepackage{amsfonts}       
\usepackage{nicefrac}       
\usepackage{microtype}      
\usepackage{lipsum}		
\usepackage{graphicx}
\usepackage{natbib}
\usepackage{doi}
\usepackage{algorithm}
\usepackage{algorithmic}

\usepackage{amsmath}
\usepackage{amssymb}
\usepackage{mathtools}
\usepackage{amsthm}
\usepackage{bbm}
\usepackage{tikz}
\usetikzlibrary{graphs,matrix,positioning}
\usetikzlibrary{bayesnet}
\usetikzlibrary{arrows}
\usetikzlibrary{shapes.geometric}
\usetikzlibrary{trees}
\usepackage[capitalize,noabbrev]{cleveref}
\usepackage[justification=centering]{caption}

\theoremstyle{plain}
\newtheorem{theorem}{Theorem}[section]

\theoremstyle{definition}

\newtheorem{assumption}[theorem]{Assumption}
\theoremstyle{remark}

\title{Regularizing Adversarial Imitation Learning Using Causal Invariance}

\author{{Ivan Ovinnikov}\\
	Department of Computer Science\\
	ETH Z\"urich\\
	Z\"urich, Switzerland \\
	\texttt{ivan.ovinnikov@inf.ethz.ch} \\
	\And
	{Joachim M. Buhmann} \\
	Department of Computer Science\\
	ETH Z\"urich\\
	Z\"urich, Switzerland \\
	\texttt{jbuhmann@inf.ethz.ch} \\
}

\renewcommand{\shorttitle}{Regularizing Adversarial Imitation Learning Using Causal Invariance}

\hypersetup{
pdftitle={A template for the arxiv style},
pdfsubject={q-bio.NC, q-bio.QM},
pdfauthor={Ivan Ovinnikov, Joachim M. Buhmann},
pdfkeywords={Causal invariance, Imitation Learning, Adversarial training},
}

\begin{document}
\maketitle

\begin{abstract}
    Imitation learning methods are used to infer a policy in a Markov decision process from a 
    dataset of expert demonstrations by minimizing a divergence measure
    between the empirical state occupancy measures of the expert and the policy. 
    The guiding signal to the policy is provided by the discriminator used 
    as part of an adversarial optimization procedure. We observe that this model is prone 
    to absorbing spurious correlations present in the expert data.
    To alleviate this issue, we propose 
    to use causal invariance as a regularization principle for adversarial training of these models.
    The regularization objective is applicable in a straightforward manner to existing 
    adversarial imitation frameworks. We demonstrate the efficacy of the 
    regularized formulation in an illustrative two-dimensional setting 
    as well as a number of high-dimensional robot locomotion benchmark tasks.
\end{abstract}


%

\section{Introduction}
The invariant causal prediction principle \cite{peters2015icp} has gained a lot of attention
in the recent years. 
Contemporary methods such as \cite{arjovsky2019irm, chang2020invariant, krueger2021out} propose 
a representation learning scheme for supervised learning problems
which aim to eliminate features which are spuriously correlated with
the target label. Various instantiations of this principle obtain 
asymptotically stable label conditionals 
across interventional settings of the data generating process. The canonical example of deep learning models absorbing such spurious 
features is the classification of cows and camels. In this example, the model learns 
the feature encoding of the background as a form of shortcut
for classifying the more complex geometry of animal shapes, 
exploiting a selection bias in the dataset. The model subsequently fails on a test set of images with permuted backgrounds.
Analogously, in reinforcement learning, it is desirable to avoid the acquisition of behaviours which would 
exploit such features. This is particularly relevant when learning from demonstrations, i.e. in the imitation learning setting.

Modern imitation learning methods \citep{ho2016generative} aim to minimize a
discrepancy measure between the a finite dataset of expert demonstrations 
and the trajectories induced by the policy trying to mimic the expert. 
The discrepancy measure is typically an instance of the family of 
$\varphi$-divergences \citep{csiszar1972class} or integral probability metrics (e.g. Wasserstein distance).
In both cases, the variational formulation of the density matching
problem is chosen for computational purposes whitch 
has been shown to have strong links to binary classification \citep{nguyen2009surrogate, sriperumbudur2009integral}, 
a fact widely used in generative adversarial network (GAN) and 
adversarial imitation methods.

In this work, we observe that the binary classifier used as discriminator in the adversarial optimization scheme is prone to exploiting the spurious correlations
present in the mixture of policy and expert trajectory data. This has multiple far-reaching implications for the resulting training procedure. 
For instance, this could lead to undesired behaviours, similar to the ones associated with 
reward hacking \cite{skalse2022defining}.
The exploitation of spurious correlations by a model typically leads to higher empirical 
performance at training time but will fail at test time.
In the context of adversarial training, an overly confident discriminator
is known to impede meaningful generator training due to a stale training signal.
This issue is typically remedied by regularizing the discriminator in various ways
\cite{gulrajani2017improved, peng2018variational}.
The problem is exacerbated by the fact that the policy will try to optimize 
the expected density ratio based on spurious features of the discriminator, 
further contributing to the covariate shift. 

To alleviate this issue, we propose to regularize the discriminator using
the invariant risk minimization principle \cite{arjovsky2019irm}, more specifically, 
the IRMv1 objective. The application of this regularization technique 
requires mild assumptions on the problem setting, which are often satisfied in practice,
and is easy to implement.
To validate our method, we perform an empirical study of the algorithm performance
in both a low-dimensional navigation setting as well as on a number of benchmark tasks from the MuJoCo suite.
We observe a consistent improvement in both settings when using the regularized version
of common adversarial imitation learning algorithms.









\section{Related work}

\paragraph{Invariance and causality in reinforcement learning} The concept of invariance has been used in a number of works in the 
reinforcement learning domain.
Invariant causal prediction has been utilized in \citep{zhang2020invariant}
to learn model invariant state abstractions in a multiple MDP setting with a shared latent space.
Invariant policy optimization \citep{sonar2021invariant} uses
the IRM games \citep{ahuja2020irmg} formulation to learn policies invariant to certain 
domain variations.
\citet{de2019causal} tackle the problem of causal confusion in 
imitation learning by making use of causal structure of demonstrations.
The issue of discriminator overfitting to task-irrelevant visual features
is addressed in \cite{zolna2021task}. Another example of using 
causal invariance is presented in \cite{bica2021invariant}.
In contrast to the methods outlined above, 
our method specifically addresses the issues with spurious correlations \emph{during} 
the process of adversarial training, which lead to discriminator degeneration.

\section{Problem setting}
We start by introducing the necessary notation and formalism to describe the problem 
setting.
\paragraph{MDP} We consider environments modelled by a \emph{Markov decision
		process} $\cM = (\cS, \cA, \cT, \mu, R)$, where $\cS$ is the state
space, $\cA$ is the action space, $\mathcal{T}$ is the family of transition distributions on
$\cS$ indexed by $\cS \times \cA$ with $p(s'|s,a)$ describing the
probability of transitioning to state $s'$ when taking action $a$ in
state $s$, $\mu$ is the initial state distribution, and
$R: \cS \times \cA \to \R$ is the reward function. A \emph{policy} $\pi$ is
a map from states $s \in \cS$ to distributions $\pi(\cdot | s)$ over
actions, with $\pi(a|s)$ being the probability of taking action $a$ in
state $s$. We denote by $\rho_E = \sum_{s_i \in \cD_E} \delta_i(s_i)$ the empirical state occupancy measure of the expert
based on a dataset of expert trajectories $\mathcal{D}_E = \{\tau_i\}_{i\leq K}$ where $\tau_i = (s_{1:T}^{(i)},a_{1:T}^{(i)})$ is a
sequence of states and actions of expert $i$ of length $T$. 
$\rho_\pi = \sum_{t \leq T} P_\mu^\pi(S_t=s, A_t=a)$ denotes the state occupancy measure induced by the policy $\pi$
over a finite horizon $T$ for initial measure $\mu$.
\paragraph{Imitation learning}%
\!\!\!\! methods aim to estimate a policy $\pi_\theta$ parameterized by 
weights $\theta$ , which mimics the expert. To achieve this goal, 
a distance or divergence measure between the empirical state occupancy measure of the expert $\rho_E$
and the induced state occupancy measure of the policy $\rho_\pi$ is minimized.
More specifically, the divergence measure is typically an instance of the class
of $\varphi$-divergences \cite{csiszar1972class}, where the choice the $\varphi$-function
corresponds to commonly used methods such as GAIL (\cite{ho2016generative}), AIRL
(\cite{fu2017learning}) or f-IRL (\cite{ni2021f}).
The adversarial imitation learning (AIL) objective is formulated as follows:
\begin{align*}
    \mathcal{L}_{AIL} = \minim{\theta} \maxim{\psi} &\mathbb{E}_{\rho_E}[\log D_\psi(s,a,s')] +\mathbb{E}_{\rho_{\pi_\theta}}[\log(1- D_\psi(s,a,s'))] - \lambda \mathcal{H}(\pi_\theta)
\end{align*}
where $D_\psi(s,a,s')$ is the discriminator parametrized by a neural network with parameters $\psi$, 
$\pi_\theta$ is the student policy and $\mathcal{H}(\pi_{\theta})$ the entropy regularization term.
\footnote{In the case of AIRL, $\psi$ denotes the joint set of structured discriminator parameters of functions $g_\xi$ and $h_\phi$}



\paragraph{Invariant causal prediction}
The principle of invariant causal prediction \cite{peters2015icp, heinzedeml2017icp} stipulates that 
for provable out-of-distribution (OOD) generalization in linear regression tasks, 
the regression coefficients must be stable across interventional settings
of the data generating process, indexed by $e \in \mathcal{E}$ where 
$\mathcal{E}$ denotes the set of datasets sampled from the data generating process.
The authors of \cite{arjovsky2019irm} extend this to nonlinear features and introduce a tractable approximation of the bi-level optimization 
required to identify invariant features $\Phi$. The derived gradient norm penalty $\mathbb{D}(w, \Phi, e) = ||\nabla_{w|w=1.0} \mathcal{L}^e(w \circ \Phi)||^2$
quantifies the violation of the normal equations to measure the optimality of a fixed linear classifier ($w=1.0$) at each setting $e$.
This leads to the following regularized formulation of the empirical risk minimization (ERM) problem where 
$\mathcal{E}_{tr} \subseteq \mathcal{E}$ is the set of training environments:
\begin{equation}
	\underset{\Phi: \mathcal{X} \to \mathcal{Y}}{\text{min}}
	\sum_{e\in\mathcal{E}_{tr}} \mathcal{L}^e(\Phi) + \lambda ||\nabla_{w|w=1.0} \mathcal{L}^e(w \circ \Phi)||^2
    \tag{IRMv1}
    \label{eq:irmv1}
\end{equation}
In the following section, we will address why this regularization is beneficial for disciminator fitting 
in adversarial training procedures.
\begin{figure*}[htp]
		\begin{minipage}{.23\linewidth}
		\centering
		\begin{tikzpicture}[scale=0.25, every node/.style={scale=0.5}]

			\node[circle, draw=blue!80, ultra thick,  ,yshift=0cm,xshift=6cm] (o1) {$\mathcal{O}_t$};
			\node[latent, xshift=1cm,yshift=0cm] (e) {$E$};
			\node[xshift=1cm,yshift=-1.5cm] (lbl) {$(a)$};
			\node[obs, xshift=2.7cm,yshift=0cm] (c) {$W$};
			\node[circle, draw=blue!80, ultra thick, xshift=4cm,yshift=1.5cm] (s1) {$s_t$};
			\node[latent, xshift=4cm,yshift=-1.5cm] (s2) {$s_{t+1}$};
			\node[circle, draw=blue!80, ultra thick, xshift=4cm,yshift=-0cm] (a1_) {$a_t$};
			\edge {e} {c}
			\edge[bend left=45] {e} {s1}
			\edge[bend right=45] {e} {s2}
			\edge {s1,a1_} {o1}
			\edge {a1_} {s2}
			\edge {c} {s1,a1_,s2}
			\edge {a1_} {s2}
			\path (s1) edge [bend left,->]  (s2) ;
			\path (c) edge [bend right,->]  (o1) ;
			

		\end{tikzpicture}

	    \label{fig:transmodel1}
		\end{minipage}
  		\begin{minipage}{.23\linewidth}
		\centering
		\begin{tikzpicture}[scale=0.25, every node/.style={scale=0.5}]

			\node[circle, draw=blue!80, ultra thick,  ,yshift=0cm,xshift=6cm] (o1) {$\mathcal{O}_t$};
			\node[latent, xshift=1cm,yshift=0cm] (e) {$E$};
			\node[xshift=1cm,yshift=-1.5cm] (lbl) {$(b)$};
			\node[obs, xshift=2.7cm,yshift=0cm] (c) {$W$};
			\node[circle, draw=blue!80, ultra thick, xshift=4cm,yshift=1.5cm] (xc) {$\mathbf{x}^{(c)}$};
			\node[circle, draw=orange, xshift=4cm,yshift=-1.5cm] (xnc) {$\mathbf{x}^{(nc)}$};
			\edge {e} {c}
			\edge[bend left=45] {e} {xc}
			\edge[bend right=45 ,draw=orange] {e} {xnc}
			\edge {xc} {o1}
			\edge {c} {xc,xnc,o1}
			\edge[draw=orange] {o1} {xnc}
			

		\end{tikzpicture}
	    \label{fig:transmodel0}
		\end{minipage}
		\begin{minipage}{.23\linewidth}
		\centering
		\begin{tikzpicture}[scale=0.25, every node/.style={scale=0.5}]

			\node[circle, draw=blue!80, ultra thick,  ,yshift=0cm,xshift=6cm] (o1) {$\mathcal{O}_t$};
			\node[latent, xshift=1cm,yshift=0cm] (e) {$E$};
			\node[xshift=1cm,yshift=-1.5cm] (lbl) {$(c)$};
			\node[obs, xshift=2.7cm,yshift=0cm] (c) {$W$};
			\node[circle, draw=blue!80, ultra thick, xshift=4cm,yshift=1.5cm] (s1) {$s_t$};
			\node[latent, draw=red, xshift=4cm,yshift=-1.5cm] (s2) {$s_{t+1}$};
			\node[obs, draw=red, xshift=4cm,yshift=-0cm] (a1_) {$a_t$};
			\edge {e} {c}
			\edge[bend left=45] {e} {s1}
			\edge[bend right=45, draw=red] {e} {s2}
			\edge {s1} {o1}
			\edge[draw=red] {a1_} {o1}
			\edge {a1_} {s2}
			\edge {c} {s1,a1_,s2}
			\edge[draw=red] {a1_} {s2}
			\path (s1) edge [bend left,->]  (s2) ;
			\path (c) edge [bend right,->]  (o1) ;
			

		\end{tikzpicture}
	    \label{fig:transmodel2}
		\end{minipage}
		\begin{minipage}{.23\linewidth}
		\centering
		\begin{tikzpicture}[scale=0.25, every node/.style={scale=0.5}]

			\node[circle, draw=blue!80, ultra thick,  ,yshift=0cm,xshift=6cm] (o1) {$\mathcal{O}_t$};
			\node[latent, xshift=1cm,yshift=0cm] (e) {$E$};
			\node[xshift=1cm,yshift=-1.5cm] (lbl) {$(d)$};
			\node[obs, xshift=2.7cm,yshift=0cm] (c) {$W$};
			\node[circle, draw=blue!80, ultra thick, xshift=4cm,yshift=1.5cm] (s1) {$s_t$};
			\node[latent, draw=orange, xshift=4cm,yshift=-1.5cm] (s2) {$s_{t+1}$};
			\node[circle, draw=blue!80, ultra thick, xshift=4cm,yshift=-0cm] (a1_) {$a_t$};
			\edge {s1} {a1_}
			\edge {e} {c}
			\edge[bend left=45] {e} {s1}
			\edge[bend right=45 ,draw=orange] {e} {s2}
			\edge {s1,a1_} {o1}
			\edge {a1_} {s2}
			\edge {c} {s1,a1_,s2}
			\edge {a1_} {s2}
			\edge[draw=orange] {o1} {s2}
			\path (s1) edge [bend left,->]  (s2) ;
			\path (c) edge [bend right,->]  (o1) ;
			

		\end{tikzpicture}
	    \label{fig:transmodel3}
		\end{minipage}

      \centering

		\caption{(a) Probabilistic graphical model of a transition under influence of the index variable $E$ and latent variable $W$. 
		The stable conditional is highlighted in blue. (b) General setting where $\mathcal{O}_t$ depends on causal $\mathbf{x}^{(c)}$ and non-causal $\mathbf{x}^{(nc)}$ features of the transition. 
		(c) Spurious correlations assuming wrong edge orientation $\mathcal{O}_t \to s_{t+1}$.
		(d) Spurious correlations assuming state-only formulation. }
		\label{fig:pgm}
\end{figure*}
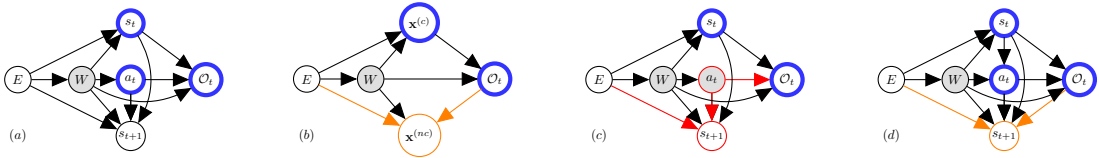
\section{Spurious correlations in adversarial imitation learning}
\label{method}
We will now describe the mechanisms by which spurious correlations lead to issues
in the process of adversarial training.
At every adversarial optimization round, a number of discriminator gradient updates is performed.
In particular, at round $k$, the discriminator is updated using the concatenated 
transition samples from the expert dataset and the policy buffer, 
$\cD_k = (\cD_E, \cD_\pi)$. Let us assume a decomposition of the feature space $\mathbf{x} \in \cX$ into two disjoint subsets, 
$\mathbf{x}^{(c)}$ and $\mathbf{x}^{(nc)}$, which denote the causal and 
spurious\footnote{Here, spuriousness is defined w.r.t. the output label} features respectively. 
By definition, spurious features do not generalize outside of the training set. 
The discriminator parametrizes the density ratio
$D(\mathbf{x}) = D(\mathbf{x}^{(c)}, \mathbf{x}^{(nc)}) = \hat{\rho}_E(\mathbf{x}^{(c)}, \mathbf{x}^{(nc)})/\hat{\rho}_\pi(\mathbf{x}^{(c)}, \mathbf{x}^{(nc)})$,
where $\hat{\rho}_E(\mathbf{x}^{(c)}, \mathbf{x}^{(nc)})$ and $\hat{\rho}_\pi(\mathbf{x}^{(c)}, \mathbf{x}^{(nc})$ denote the density estimators 
of the expert and policy respectively.
Suppose in the cow-camel classification example, the camera operator is given 
agency to move the camera to frame different parts of the scene. The guiding 
signal is provided by the negative log likelihood ratio of features 
present in subsets of the cow and camel images, meaning deviations from
unity ratio are penalized. If the ratio is easier to estimate using the $\mathbf{x}^{(nc)}$
variable of the joint distribution $\hat{p}(\mathbf{x}^{(c)}, \mathbf{x}^{(nc)}$), the camera
might end up focusing on parts of the scene which do not contain objects of interest.
Analogously, in the imitation learning case, 
the policy occupancy measure will converge to a part of the state space, 
which might not describe meaningful behaviours.

\label{sec:spur_ail}
\paragraph{Spurious correlations in model of transitions} Figure 1
describes our setting from a probabilistic graphical model point of view.
We consider various settings in which a non-causal information path corresponding to spurious correlations 
is formed in the structural causal model (SCM) of transitions (Fig. 1a). 
Fig. 1b illustrates the most general setting where an arbitrary transition input $(s,a,s')$ is partitioned into 
the causal transition feature components $\mathbf{x}^{(c)}$ and 
$\mathbf{x}^{(nc)}$ : $(s,a,s') = (\mathbf{x}^{(c)},\mathbf{x}^{(nc)})$, 
whereby conditioning on the $\mathbf{x}^{(nc)}$ collider introduces 
a spurious correlation path.
In Fig. 1c we can observe the scenario where we do not 
condition the discriminator $D(s)$ on the action. By conditioning 
on the collider node $s_{t+1}$ and not observing 
the action node $a_t$, a path is formed between the setting index $E$ and the optimality 
variable $\mathcal{O}_t$, resulting in the violation of their conditional independence
relationship. A third scenario 
can be observed in Fig. 1d. This scenario requires the assumption that the orientation 
of the edge from node $\mathcal{O}_t$ to node $s_{t+1}$ is temporally causal, meaning
that the optimality of a state at time $t$ is a causal parent of the next state.
In this case, observing the collider node $s_{t+1}$ implies the following 
conditional independence relationship: $E \not\!\perp\!\!\!\perp O_t | s_{t+1}$.

In order to apply this intuition in the desired context, we must make the following assumption which has implications  
on the necessary data and training procedure specifics.
\begin{assumption}
\emph{The data samples in the discriminator training tuple at round $k$ $\cD_k$ stem from different \emph{settings} $e \in \mathcal{E}_{tr}$.}
\label{assu1}
\end{assumption}

\begin{algorithm}[tb]
   \caption{Causally invariant adversarial imitation learning (CIAIL)}
   \label{algo:ciil}
\begin{algorithmic}
   \STATE {\bfseries Input:} Expert trajectories $\mathcal{D}^e_E$ from settings $e\in \mathcal{E}$
   \STATE Initialize \emph{actor-critic} $(\pi_\theta, V_\vartheta)$ or \emph{soft actor-critic} $(\pi_\theta, Q_\varsigma^{(j)}, V_\vartheta)$ 
   and discriminator $D_{\psi}$

   \FOR{$t=1$ {\bfseries to} $N_{rounds}$}
   \STATE Collect trajectory buffer $\mathcal{D}_\pi = \{\tau_i\}_{i \leq |\mathcal{D}_\pi|}$ by executing the policy $\pi_\theta$\;
   \STATE Update $D_{\psi}(s,a)$ via binary logistic regression by maximizing $\mathcal{L}(\psi,e)$  using tuple $\cD_t = (\cD_E,\cD_\pi)$:
			\begin{align*} 
            \mathcal{L}(\psi; e) = \mathcal{L}_{\text{BCE}}(\psi;e)+  \lambda ||\nabla_{\omega|\omega=1.0} \mathcal{L}_{\text{BCE}}(\psi; e)||^2
            \end{align*}

   \STATE Compute $\log D_\psi(s,a,s') \;\forall (s,a,s')\in\cD$
   \STATE 1. (On-policy CIAIL): Update $(\pi_\theta, V_\vartheta)$ using a constrained policy gradient method (e.g. PPO) using $r_\psi$ as reward
   \STATE 2. (Off-policy CIAIL): Update $(\pi_\theta, Q_\varsigma^{(j)}, V_\vartheta)$ using an off-policy methods (e.g. SAC) using $r_\psi$ as reward 

   \ENDFOR
\end{algorithmic}
\end{algorithm}
\begin{table*}[htp]
\small
	\caption{Policy rollout results using ground truth reward for 2d navigation environment (\texttt{MovePoint}) with varying regularization strength and number of discriminator updates. 
    Expert reference: -23665.025$_{\pm 2264.521}$}
	\label{table_2denv}
	\centering
\begin{tabular}{llllll}
\toprule
$n_{updates}$ &                   irm: $\lambda=0.01$ &                    irm: $\lambda=0.1$ &                    irm: $\lambda=1.0$&                  irm: $\lambda=10.0$&                        erm \\
\midrule
GAIL & & & & & \\        
\midrule
1  &   -24747.54$_{\pm4386.43}$ &    -25702.6$_{\pm3534.82}$ &   -24732.45$_{\pm3270.27}$ &   \textbf{-22836.3$_{\pm3240.09}$} &    -26703.01$_{\pm4607.2}$ \\
2  &   -27169.45$_{\pm4005.41}$ &   -28277.35$_{\pm4534.81}$ &   -32413.52$_{\pm4023.55}$ &  -27741.34$_{\pm4692.81}$ &    \textbf{-22320.1$_{\pm2181.62}$} \\
5  &   \textbf{-25339.01$_{\pm3560.01}$} &   -28124.89$_{\pm5341.83}$ &   -33132.36$_{\pm6205.38}$ &  -28396.54$_{\pm2183.32}$ &   -30267.07$_{\pm4701.83}$ \\
10 &    -36217.57$_{\pm5066.6}$ &   -34263.65$_{\pm6112.41}$ &   \textbf{-29385.81$_{\pm3815.97}$} &  -34222.82$_{\pm5145.31}$ &   -33012.71$_{\pm5759.67}$ \\
\midrule
AIRL & & & & & \\        
\midrule
1  &   \textbf{-29958.88$_{\pm4681.55}$} &    -34682.4$_{\pm4199.57}$ &   -33810.05$_{\pm5590.15}$ &  -30720.46$_{\pm4385.96}$ &   -30003.58$_{\pm2758.92}$ \\
2  &   -42563.59$_{\pm5112.57}$ &    \textbf{-30877.76$_{\pm4797.0}$} &   -31796.58$_{\pm4120.83}$ &  -45376.83$_{\pm9748.12}$ &   -34177.21$_{\pm6388.29}$ \\
5  &   -34297.02$_{\pm6927.63}$ &   -39084.49$_{\pm7758.24}$ &   \textbf{-33692.61$_{\pm6301.28}$} &  -43255.18$_{\pm6108.74}$ &    -42262.5$_{\pm6552.44}$ \\
10 &   -33756.08$_{\pm5623.63}$ &   -40828.81$_{\pm6786.65}$ &    -38408.9$_{\pm6991.46}$ &  \textbf{-30690.02$_{\pm4484.04}$} &   -35218.11$_{\pm3468.91}$ \\
\bottomrule
\end{tabular}
\end{table*}
The assumption is motivated by the fact that in the \ref{eq:irmv1} formulation, no 
explicit environment specification is necessary to perform the optimization and obtain invariant features.
For the problem we are considering, this assumption is satisfied in two cases. The first case necessitates a \emph{varied} set of 
expert demonstrations $\cD_E^e$ where $e$ denotes the setting index. This scenario 
is quite common as the expert demonstrations from multiple sources are typically 
pooled into one dataset.
The second case corresponds to a setting where the policy set of transitions $\cD_\pi$
contains transitions gathered over multiple policy optimization episodes. This setting  
corresponds to off-policy reinforcement learning methods such as Soft Actor-Critic (SAC) \cite{haarnoja2018soft},
where the replay buffer contains rollouts of policies from previous optimization rounds.
\paragraph{Algorithm}
We outline the proposed algorithm in \ref{algo:ciil}. The algorithm  introduces two novel aspects to the adversarial imitation learning pipeline. The first is a straightforward application of the IRMv1 penalty 
to the discriminator binary cross-entropy loss. This can be applied to both the \emph{on-policy} and the \emph{off-policy} 
formulations of the algorithm. The on-policy formulation utilizes the Proximal Policy Optimization (PPO) \citep{schulman2017proximal} algorithm for policy training.
The introduction of an off-policy algorithm, Soft Actor-Critic (SAC) \citep{haarnoja2018soft} is the second addition. The 
use of an off-policy algorithm has previously been explored in \cite{kostrikov2018discriminator} for the purposes
of sample efficiency. In our case, the off-policy formulation is one of the scenarios which satisfies \ref{assu1}.
\section{Experiments}
In order to evaluate our method empirically, 
we propose to conduct two different experiments. 
In both experiments, we compare the performance of the proposed
regularization applied to two well-established adversarial
imitation learning baselines: GAIL \cite{ho2016generative}
and AIRL \cite{fu2017learning}.

\subsection{2d goal navigation}
The first task consists of a simple two-dimensional goal navigation problem (\texttt{MoveP-v0}) defined on 
states given by the concatenation of Cartesian
coordinates of the agent and the target and imbued with a discrete action space corresponding
to the movement directions. 
The 10 expert trajectories used in training are obtained by sampling a policy trained 
on the ground truth reward defined as the Euclidean distance to target.
In order to simulate the fact that experts stem from different settings, 
we introduce an intermediate goal which varies across the experts. 
Here, we limit our evaluation to the on-policy version of algorithm \ref{algo:ciil}.
The regularization coefficient $\lambda$ is varied in the range
$\lambda \in \{0.01,0.1,1.0,10.0\}$ and the number of discriminator 
updates is varied in the range $n \in \{1,2,5,10\}$. 
The results are summarized in Table \ref{table_2denv}. We can 
observe that applying the causal invariance penalty has a consistent positive effect 
when evaluating the rollout performance of the policies.

\subsection{MuJoCo robot locomotion}
The second setting is a subset of MuJoCo robot locomotion tasks.
Here, we evaluate both the on-policy and the off-policy formulations of the presented algorithm.
In Table 2, we can observe that for both policy learning algorithms,
regularizing the discriminator significantly improved the cumulative ground truth reward 
metric obtained by rolling out the learned policies.
In particular, we observe a dramatic improvement for the case where the off-policy
algorithm (SAC) is used for policy optimization, which validates our assumption \ref{assu1}.
Our algorithm also favorably compares to an existing gradient penalty regularization method which 
is based on the convex combination of inputs (mixup) \cite{carratino2020mixup} denoted by the $GP$ suffix in Table 2.

\begin{table*}[htp]
\scriptsize	
\centering{
	\caption{Policy rollout results using ground truth reward for MuJoCo environments.
     The \texttt{GP} suffix corresponds to the gradient penalty regularization with regularization coefficient $\lambda_{GP}=50.0$ 
     and \texttt{IRM} suffix to the IRM regularization with coefficient $\lambda_{IRM}=50.0$. The algorithms were trained
     on 10 expert trajectories with the following recorded rollout performance: Ant-v3: 4303.532$_{\pm1553.060}$, 
     HalfCheetah-v3: 9018.685$_{\pm125.446}$, Hopper-v3: 1709.923$_{\pm859.010}$, Walker2d-v3: 3984.531$_{\pm64.259}$ }}
	\label{table_mujoco_gp}
	\centering
\begin{tabular}{lllllll}
\toprule
Environment & GAIL-ERM & GAIL-GP & GAIL-IRM & AIRL-ERM & AIRL-GP & AIRL-IRM\\
\midrule
SAC & & & & &  \\
\midrule
Ant-v3 & 2163.28$_{\pm1835.18}$ & 3292.43$_{\pm1365.91}$ & \textbf{4291.28$_{\pm1243.44}$} & 361.35$_{\pm218.897}$ & 9.69$_{\pm3.89}$&\textbf{2010.191$_{\pm2170.729}$} \\ 
HalfCheetah-v3 & 941.69$_{\pm382.93}$ & 1983.39$_{\pm382.93}$  & \textbf{2352.82$_{\pm733.15}$} & 2666.90$_{\pm515.74}$ & 2849.03$_{\pm367.74}$ &\textbf{3450.245$_{\pm1465.968}$} \\
Hopper-v3 & 3079.70$_{\pm951.30}$ & 2819.37$_{\pm983.72}$ &\textbf{3315.53 $_{\pm956.21}$} & 3581.49$_{\pm39.04}$ & 728.96$_{\pm324.067}$&\textbf{3770.767$_{\pm61.337}$} \\
Walker2d-v3 &3128.51$_{\pm1452.83}$ & 3076.22$_{\pm1275.81}$ &\textbf{3705.04$_{\pm1068.60}$} & 2355.69$_{\pm646.41}$ & 1538.38$_{\pm1070.435}$ & \textbf{4213.480$_{\pm58.596}$}\\
\midrule
PPO & & & & &  \\
\midrule
Ant-v3 & 18.483$_{\pm12.318}$ & -28.1$_{\pm100.89}$ &\textbf{26.326$_{\pm20.265}$} & 48.650423$_{\pm9.492}$ & 8.43$_{\pm11.92}$ &\textbf{73.427$_{\pm21.424}$} \\ 
HalfCheetah-v3 & 2577.173	$_{\pm1324.064}$& \textbf{4117.92$_{\pm1214.57}$} &2788.065$_{\pm1015.200}$ & 571.990$_{\pm223.634}$& 52.01$_{\pm137.43}$&\textbf{976.267 $_{\pm1365.386}$}\\
Hopper-v3 & 2859.980$_{\pm1114.946}$ & 2708.60$_{\pm976.71}$&\textbf{3173.536 $_{\pm923.694}$} &170.987$_{\pm60.603}$ & 45.01 $_{\pm40.98}$ &\textbf{1421.980 $_{\pm1523.009}$} \\
Walker2d-v3 & 2648.653	$_{\pm1128.649}$ & 1945.99$_{\pm781.85}$ &\textbf{3443.572$_{\pm1032.796}$} & 24.773$_{\pm5.793}$ & 2.91$_{\pm3.74}$ & 
\textbf{1361.887 $_{\pm1642.685}$}\\
\bottomrule
\end{tabular}
\end{table*}

\section{Discussion}
In this work, we have presented a novel algorithm which introduces a causal invariance regularization objective to adversarial imitation learning.
We have observed its efficacy in a number of settings and described scenarios which benefits from its application.
Future work includes extending these preliminary results to the image domain and a more in depth comparison to
existing regularization techniques some of which have recently been interpreted through a causal lens.
While the emprical evaluation seems to indicate a strong benefit of the method, a more thorough
theoretical analysis of the distribution shift of the discriminator input would be beneficial.
Furthermore, a stronger link between spurious correlations and reward hacking behaviours should be established.

\nocite{langley00}

\bibliography{refs}

\begin{thebibliography}{24}
\providecommand{\natexlab}[1]{#1}
\providecommand{\url}[1]{\texttt{#1}}
\expandafter\ifx\csname urlstyle\endcsname\relax
  \providecommand{\doi}[1]{doi: #1}\else
  \providecommand{\doi}{doi: \begingroup \urlstyle{rm}\Url}\fi

\bibitem[Peters et~al.(2015)Peters, B{\"u}hlmann, and
  Meinshausen]{peters2015icp}
Jonas Peters, Peter B{\"u}hlmann, and Nicolai Meinshausen.
\newblock Causal inference using invariant prediction: identification and
  confidence intervals.
\newblock \emph{arXiv preprint arXiv:1501.01332}, 2015.

\bibitem[Arjovsky et~al.(2019)Arjovsky, Bottou, Gulrajani, and
  Lopez-Paz]{arjovsky2019irm}
Martin Arjovsky, Léon Bottou, Ishaan Gulrajani, and David Lopez-Paz.
\newblock Invariant risk minimization, 2019.

\bibitem[Chang et~al.(2020)Chang, Zhang, Yu, and Jaakkola]{chang2020invariant}
Shiyu Chang, Yang Zhang, Mo~Yu, and Tommi Jaakkola.
\newblock Invariant rationalization.
\newblock In \emph{International Conference on Machine Learning}, pages
  1448--1458. PMLR, 2020.

\bibitem[Krueger et~al.(2021)Krueger, Caballero, Jacobsen, Zhang, Binas, Zhang,
  Le~Priol, and Courville]{krueger2021out}
David Krueger, Ethan Caballero, Joern-Henrik Jacobsen, Amy Zhang, Jonathan
  Binas, Dinghuai Zhang, Remi Le~Priol, and Aaron Courville.
\newblock Out-of-distribution generalization via risk extrapolation (rex).
\newblock In \emph{International Conference on Machine Learning}, pages
  5815--5826. PMLR, 2021.

\bibitem[Ho and Ermon(2016)]{ho2016generative}
Jonathan Ho and Stefano Ermon.
\newblock Generative adversarial imitation learning.
\newblock \emph{arXiv preprint arXiv:1606.03476}, 2016.

\bibitem[Csisz{\'a}r(1972)]{csiszar1972class}
Imre Csisz{\'a}r.
\newblock A class of measures of informativity of observation channels.
\newblock \emph{Periodica Mathematica Hungarica}, 2\penalty0 (1-4):\penalty0
  191--213, 1972.

\bibitem[Nguyen et~al.(2009)Nguyen, Wainwright, and
  Jordan]{nguyen2009surrogate}
XuanLong Nguyen, Martin~J Wainwright, and Michael~I Jordan.
\newblock On surrogate loss functions and f-divergences.
\newblock 2009.

\bibitem[Sriperumbudur et~al.(2009)Sriperumbudur, Fukumizu, Gretton,
  Sch{\"o}lkopf, and Lanckriet]{sriperumbudur2009integral}
Bharath~K Sriperumbudur, Kenji Fukumizu, Arthur Gretton, Bernhard
  Sch{\"o}lkopf, and Gert~RG Lanckriet.
\newblock On integral probability metrics,$\backslash$phi-divergences and
  binary classification.
\newblock \emph{arXiv preprint arXiv:0901.2698}, 2009.

\bibitem[Skalse et~al.(2022)Skalse, Howe, Krasheninnikov, and
  Krueger]{skalse2022defining}
Joar Skalse, Nikolaus H.~R. Howe, Dmitrii Krasheninnikov, and David Krueger.
\newblock Defining and characterizing reward hacking, 2022.

\bibitem[Gulrajani et~al.(2017)Gulrajani, Ahmed, Arjovsky, Dumoulin, and
  Courville]{gulrajani2017improved}
Ishaan Gulrajani, Faruk Ahmed, Martin Arjovsky, Vincent Dumoulin, and Aaron~C
  Courville.
\newblock Improved training of wasserstein gans.
\newblock \emph{Advances in neural information processing systems}, 30, 2017.

\bibitem[Peng et~al.(2018)Peng, Kanazawa, Toyer, Abbeel, and
  Levine]{peng2018variational}
Xue~Bin Peng, Angjoo Kanazawa, Sam Toyer, Pieter Abbeel, and Sergey Levine.
\newblock Variational discriminator bottleneck: Improving imitation learning,
  inverse rl, and gans by constraining information flow.
\newblock \emph{arXiv preprint arXiv:1810.00821}, 2018.

\bibitem[Zhang et~al.(2020)Zhang, Lyle, Sodhani, Filos, Kwiatkowska, Pineau,
  Gal, and Precup]{zhang2020invariant}
Amy Zhang, Clare Lyle, Shagun Sodhani, Angelos Filos, Marta Kwiatkowska, Joelle
  Pineau, Yarin Gal, and Doina Precup.
\newblock Invariant causal prediction for block mdps.
\newblock In \emph{International Conference on Machine Learning}, pages
  11214--11224. PMLR, 2020.

\bibitem[Sonar et~al.(2021)Sonar, Pacelli, and Majumdar]{sonar2021invariant}
Anoopkumar Sonar, Vincent Pacelli, and Anirudha Majumdar.
\newblock Invariant policy optimization: Towards stronger generalization in
  reinforcement learning.
\newblock In \emph{Learning for Dynamics and Control}, pages 21--33. PMLR,
  2021.

\bibitem[Ahuja et~al.(2020)Ahuja, Shanmugam, Varshney, and
  Dhurandhar]{ahuja2020irmg}
Kartik Ahuja, Karthikeyan Shanmugam, Kush Varshney, and Amit Dhurandhar.
\newblock Invariant risk minimization games.
\newblock \emph{arXiv preprint arXiv:2002.04692}, 2020.

\bibitem[de~Haan et~al.(2019)de~Haan, Jayaraman, and Levine]{de2019causal}
Pim de~Haan, Dinesh Jayaraman, and Sergey Levine.
\newblock Causal confusion in imitation learning.
\newblock \emph{arXiv preprint arXiv:1905.11979}, 2019.

\bibitem[Zolna et~al.(2021)Zolna, Reed, Novikov, Colmenarejo, Budden, Cabi,
  Denil, de~Freitas, and Wang]{zolna2021task}
Konrad Zolna, Scott Reed, Alexander Novikov, Sergio~Gomez Colmenarejo, David
  Budden, Serkan Cabi, Misha Denil, Nando de~Freitas, and Ziyu Wang.
\newblock Task-relevant adversarial imitation learning.
\newblock In \emph{Conference on Robot Learning}, pages 247--263. PMLR, 2021.

\bibitem[Bica et~al.(2021)Bica, Jarrett, and van~der Schaar]{bica2021invariant}
Ioana Bica, Daniel Jarrett, and Mihaela van~der Schaar.
\newblock Invariant causal imitation learning for generalizable policies.
\newblock \emph{Advances in Neural Information Processing Systems},
  34:\penalty0 3952--3964, 2021.

\bibitem[Fu et~al.(2017)Fu, Luo, and Levine]{fu2017learning}
Justin Fu, Katie Luo, and Sergey Levine.
\newblock Learning robust rewards with adversarial inverse reinforcement
  learning.
\newblock \emph{arXiv preprint arXiv:1710.11248}, 2017.

\bibitem[Ni et~al.(2021)Ni, Sikchi, Wang, Gupta, Lee, and Eysenbach]{ni2021f}
Tianwei Ni, Harshit Sikchi, Yufei Wang, Tejus Gupta, Lisa Lee, and Ben
  Eysenbach.
\newblock f-irl: Inverse reinforcement learning via state marginal matching.
\newblock In \emph{Conference on Robot Learning}, pages 529--551. PMLR, 2021.

\bibitem[Heinze-Deml et~al.(2017)Heinze-Deml, Peters, and
  Meinshausen]{heinzedeml2017icp}
Christina Heinze-Deml, Jonas Peters, and Nicolai Meinshausen.
\newblock Invariant causal prediction for nonlinear models, 2017.

\bibitem[Haarnoja et~al.(2018)Haarnoja, Zhou, Hartikainen, Tucker, Ha, Tan,
  Kumar, Zhu, Gupta, Abbeel, et~al.]{haarnoja2018soft}
Tuomas Haarnoja, Aurick Zhou, Kristian Hartikainen, George Tucker, Sehoon Ha,
  Jie Tan, Vikash Kumar, Henry Zhu, Abhishek Gupta, Pieter Abbeel, et~al.
\newblock Soft actor-critic algorithms and applications.
\newblock \emph{arXiv preprint arXiv:1812.05905}, 2018.

\bibitem[Schulman et~al.(2017)Schulman, Wolski, Dhariwal, Radford, and
  Klimov]{schulman2017proximal}
John Schulman, Filip Wolski, Prafulla Dhariwal, Alec Radford, and Oleg Klimov.
\newblock Proximal policy optimization algorithms.
\newblock \emph{arXiv preprint arXiv:1707.06347}, 2017.

\bibitem[Kostrikov et~al.(2018)Kostrikov, Agrawal, Dwibedi, Levine, and
  Tompson]{kostrikov2018discriminator}
Ilya Kostrikov, Kumar~Krishna Agrawal, Debidatta Dwibedi, Sergey Levine, and
  Jonathan Tompson.
\newblock Discriminator-actor-critic: Addressing sample inefficiency and reward
  bias in adversarial imitation learning.
\newblock \emph{arXiv preprint arXiv:1809.02925}, 2018.

\bibitem[Carratino et~al.(2020)Carratino, Ciss{\'e}, Jenatton, and
  Vert]{carratino2020mixup}
Luigi Carratino, Moustapha Ciss{\'e}, Rodolphe Jenatton, and Jean-Philippe
  Vert.
\newblock On mixup regularization.
\newblock \emph{arXiv preprint arXiv:2006.06049}, 2020.

\end{thebibliography}

\bibliographystyle{unsrtnat}






\end{document}